\documentclass{elsarticle}
\usepackage{graphicx}
\usepackage{amssymb}

\begin{document}
\begin{frontmatter}

\title{A probabilistic estimation and prediction technique for dynamic continuous social science models: The evolution of the attitude of the Basque Country population towards ETA as a case study}

\author[l2]{Juan-Carlos Cort\'es}
\ead{jccortes@imm.upv.es}
\author[l1]{Francisco-J. Santonja} 
\ead{francisco.santonja@uv.es}
\author[l2]{Ana-C. Tarazona}  
\ead{actarazona@asic.upv.es}
\author[l2]{Rafael-J. Villanueva\corref{cor1}}
\ead{rjvillan@imm.upv.es}
\author[l3]{Javier Villanueva-Oller}
\ead{jvillanueva@pdi.ucm.es}

\cortext[cor1]{Corresponding Author.}

\address[l2]{Instituto Universitario de Matem\'{a}tica Multidisciplinar, Universitat Polit\`{e}cnica de Val\`{e}ncia, Valencia (Spain)}
\address[l1]{Departamento de Estad\'{\i}stica e Investigaci\'{o}n Operativa, Universitat de Val\'encia, Valencia (Spain)}
\address[l3]{Centro de Estudios Superiores Felipe II, Aranjuez, Madrid (Spain)}

\begin{abstract}
In this paper, we present a computational technique to deal with uncertainty in dynamic continuous models in Social Sciences. Considering data from surveys, the method consists of determining the probability distribution of the survey output and this allows to sample data and fit the model to the sampled data using a goodness-of-fit criterion based on the $\chi^2$-test. Taking the fitted parameters non-rejected by the $\chi^2$-test, substituting them into the model and computing their outputs, we build $95\%$ confidence intervals in each time instant capturing uncertainty of the survey data (probabilistic estimation). Using the same set of obtained model parameters, we also provide a prediction over the next few years with $95\%$ confidence intervals (probabilistic prediction). 
This technique is applied to a dynamic social model describing the evolution of the attitude of the Basque Country population towards the revolutionary organization ETA.
\end{abstract}


\end{frontmatter}

\section{Introduccion}
Uncertainty quantification in dynamic continuous models is an emerging area \cite{maitre}. Because of the numerous complex factors that usually involve social behavior, it is particularly appropriate the consideration of randomness in this kind of models. In practice, the introduction of randomness in continuous models can be done using different approaches. Stochastic differential  equations of It\^{o}-type consider uncertainty through a stochastic process called white noise, i.e., the derivative of a Wiener process. As a consequence, this approach limitates the introduction of uncertainty to a gaussian process whose sample trajectories are somewhat irregular since they are nowhere differentiable. A more convenient approach in social modelling is to permit that input parameters can become random variables and/or stochastic processes and, therefore can follow other type of probability distributions apart from gaussian. This approach leads to continuous models usually referred to as random differential equations (r.d.e.'s). In dealing with r.d.e.'s, generalized Polynomial Chaos (gPC) is likely one of the most fruitful methods \cite{Xiu, Spanos}.

Most of  the existing methods and techniques, start with the assumption that the model parameters follow a known standard probability distribution. In general, setting the probability distribution of the model parameters, standard or empirical, is a crucial and difficult task currently under study which is required for model uncertainty approaches.

Also, the computation is an important issue in dealing with uncertainty. For instance, gPC technique may not be affordable when the number of model parameters with uncertainty increases, or the interval where the mean and the standard deviation are valid may be very short \cite{Benito}. It may turn these techniques inappropriate for modelling real problems. 

On the other hand, if we consider that no information is available for setting the model parameters probabilistic distribution, techniques as bootstapping \cite{almu} or bayesian \cite{bay} are other useful approaches. Related to these statistical techniques, in this paper we propose a computational approach where the data, retrieved from surveys, play a fundamental role to introduce the uncertainty, in estimation and prediction, from the very beginning. This probabilistic approach is applied to a model describing the evolution of the attitude of the Basque population towards the revolutionary organization ETA \cite{ETA} presented in \cite{NOS}. 

The paper is organized as follows. In Section 2, we summarize the model building described in \cite{NOS}. In Section 3 we propose a technique which will allow us to obtain a set of model parameters that provide $95\%$ confidence intervals for each time instant such that the data uncertainty is captured. We will call this technique \textit{probabilistic estimation}. With the set of parameters obtained in Section 3, in Section 4 we obtain a \textit{probabilistic prediction} of the attitude towards ETA of the people of the Basque Country over the next four years. In Section 5, we discuss the results and present the conclusion.

\section{Model building}
In \cite{NOS}, the deterministic mathematical model was introduced, and series data were retrieved from the Euskobarometro of November 2012 on the attitude of the Basque Country population towards ETA \cite[Table 20]{eusko}. The eight types of attitudes towards ETA that appear in the Euskobarometro (Total support; Justification with criticism; Goals yes / Means no; Before yes / Not now; Indifferent; ETA scares; Total rejection; No answer) were simplified to only three (Support; Rejection; Abstention) and allowed us to divide Basque Country population into the following three subpopulations, time $t$ in years (see \cite{NOS} for more details):

\begin{itemize}
\item \textbf{Supporters.} $A_1(t)$, the percentage of people in the Basque Country which have an attitude of support towards ETA at time instant $t$,
\item \textbf{Rejectors.} $A_2(t)$ is the percentage of people in the Basque Country which have an attitude of rejection towards ETA at time $t$, 
\item \textbf{Abstentionists.} $A_3(t)$   is the percentage of people in the Basque Country whose attitude towards ETA is not defined (indifferent), abstain or simply they do not want to declare their opinion, at time $t$.
\end{itemize}

\begin{figure}[h]
  \centering
  \includegraphics[scale=0.25]{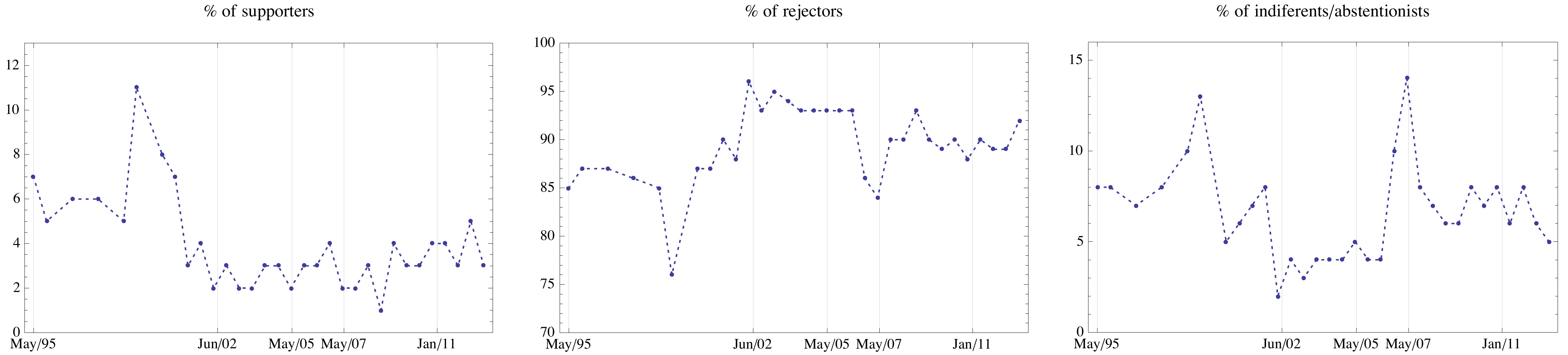}
  \caption{Percentage of Basque Country population with an attitude of support, rejection or abstention towards ETA since May 1995 until Nov 2012. Vertical lines correspond to remarkable dates: in Jun 2002 the Law of Political Parties was passed and left-wing nationalist political parties were outlawed because their proven relation with ETA; in May 2005 the Spanish Parliament approved the possibility the Government to support dialogue with ETA; in May 2007 left-wing nationalist parties could present candidates again; in Jan 2011 ETA announced a permanent cease-fire. Observe that large jumps in the Rejection population correspond to large jumps in the Abstention population, in the opposite sense. Supporting population remains with minor variations since the Law of Political Parties passed.}
\label{datos}
\end{figure}

Data grouped in these three groups appear, in percentages, in Figure \ref{datos} from May 1995 until Nov 2012. In May 2005 the Spanish Parliament approved the possibility the Government to support dialogue with ETA what has been considered as a substantial change in the anti-terrorist policy. This policy is still in force and it justifies we choose this time instant as our model initial condition. In Table \ref{TABLA1} we present the figures in percentages of each subpopulation from May 2005 to Nov 2012.

\begin{table}[h]
\centering
\begin{tabular}{|l|c|c|c|}
\hline
  Survey date  & Support (\%) & Rejection (\%) & Abstention (\%) \\ 
\hline
May 05	&	2	&	93	&	5	\\
Nov 05	&	3	&	93	&	4	\\
May 06	&	3	&	93	&	4	\\
Nov 06	&	4	&	86	&	10	\\
May 07	&	2	&	84	&	14	\\
Nov 07	&	2	&	90	&	8	\\
May 08	&	3	&	90	&	7	\\
Nov 08	&	1	&	93	&	6	\\
May 09	&	4	&	90	&	6	\\
Nov 09	&	3	&	89	&	8	\\
May 10	&	3	&	90	&	7	\\
Nov 10	&	4	&	88	&	8	\\
May 11	&	4	&	90	&	6	\\
Nov 11	&	3	&	89	&	8	\\
May 12	&	5	&	89	&	6	\\
Nov 12	&	3	&	92	&	5	\\
 \hline 
\end{tabular} 
\caption{Percentage of people in the Basque Country with respect to their attitude towards ETA from May 2005 to Nov 2012.}
\label{TABLA1} 
\end{table}

Then, as we show in \cite{NOS}, the following system of nonlinear differential equations describes the evolution of attitudes towards ETA in the Basque Country over time:

\begin{eqnarray}
A'_1(t) = &  \beta_{21} A_2(t) A_1(t) - \beta_{12} A_1(t) A_2(t) + \beta_{31} A_3(t) A_1(t) - \beta_{13} A_1(t) A_3(t), \nonumber \\
A'_2(t) = &  \beta_{12} A_1(t) A_2(t) - \beta_{21} A_2(t) A_1(t) + \beta_{32} A_3(t) A_2(t) - \beta_{23} A_2(t) A_3(t), \nonumber \\
A'_3(t) = &  \beta_{13} A_1(t) A_3(t) - \beta_{31} A_3(t) A_1(t) + \beta_{23} A_2(t) A_3(t) - \beta_{32} A_3(t) A_2(t). \nonumber
\end{eqnarray} 

Taking $\gamma_{12} = \beta_{12}  - \beta_{21}$, $\gamma_{13} = \beta_{13}  - \beta_{31}$ and $\gamma_{23} = \beta_{23}  - \beta_{32}$, the above system can be simplified as follows

\begin{eqnarray}
A'_1(t) = &  -\gamma_{12} A_2(t) A_1(t) - \gamma_{13} A_3(t) A_1(t), \label{eq1} \\
A'_2(t) = &   \gamma_{12} A_2(t) A_1(t) - \gamma_{23} A_3(t) A_2(t), \\
A'_3(t) = &   \gamma_{13} A_3(t) A_1(t) + \gamma_{23} A_3(t) A_2(t). \label{eq3}                           
\end{eqnarray} 

\begin{figure}[h]
 \begin{center}
  \includegraphics[scale=0.7]{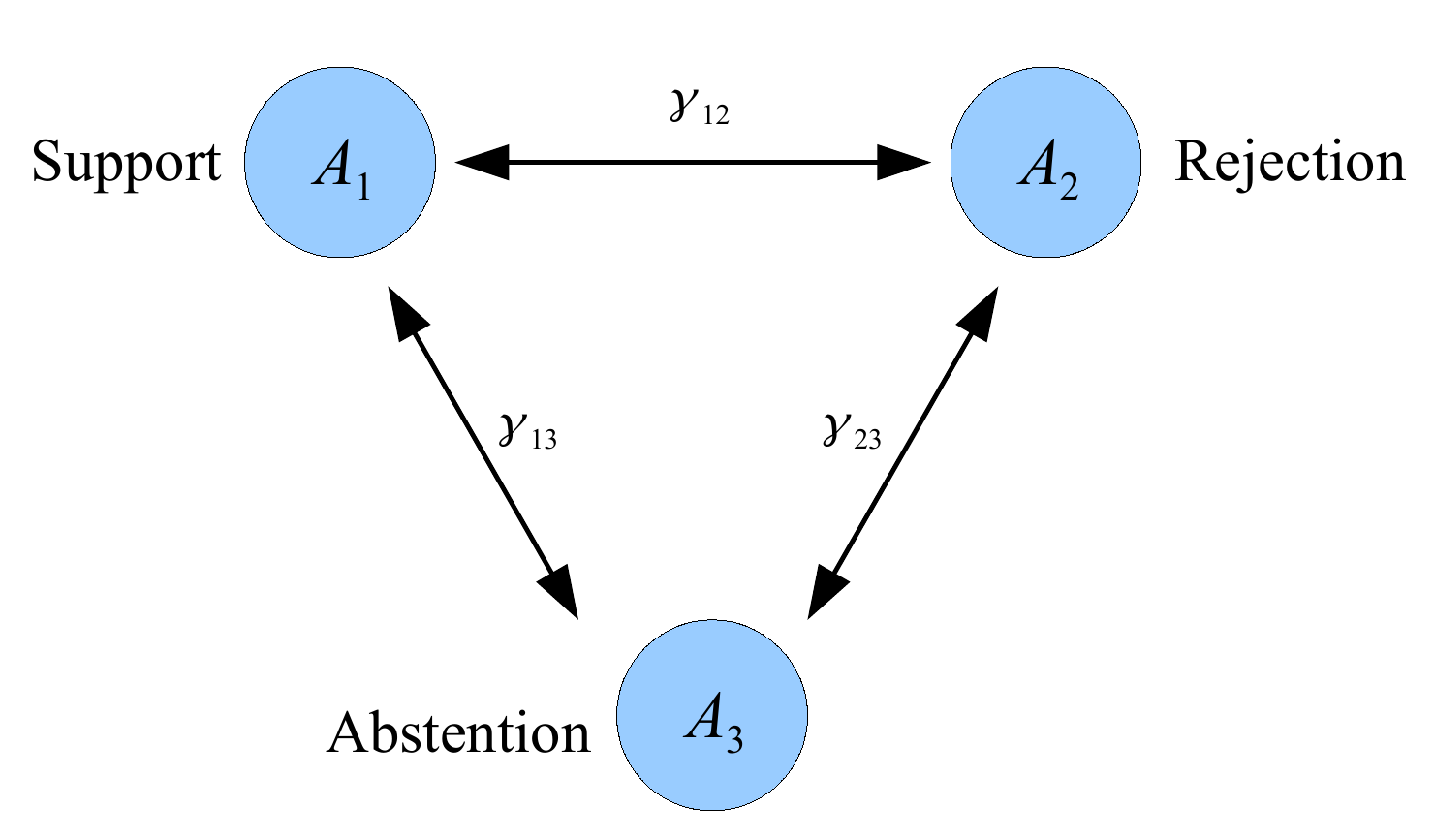}\\
  \caption{Graph depicting the model (\ref{eq1})-(\ref{eq3}). The circles are the subpopulations and the arrows represent the flow of people who change their attitude towards ETA over the time.}\label{Modelo}
\end{center}
\end{figure}

Note that if $\gamma_{ij} > 0$ the net movement of individuals is from $A_i$ to $A_j$. The above system of differential equations can be represented by the diagram of Figure \ref{Modelo}. See \cite{NOS} for more details.
 
\section{Probabilistic estimation: A computational technique to determine the empirical probabilistic distribution of model parameters}

\subsection{Data}
Data in Table \ref{TABLA1} correspond to the mean percentage obtained from the Euskobarometro surveys since May 2005 to Nov 2012 \cite[Table 20]{eusko}. In the technical specifications of each survey we can see sample sizes of $1800$ and $1200$ interviews (see column 3 in Table \ref{TABLA2}). 

Taking into account the sample is not the same for each survey, let us assume that the survey outputs are independent. For each one of the $16$ available surveys, let us denote by $X^j=(X_1^j,X_2^j,X_3^j)$, $0\leq X_i^j\leq n_j$, $i=1,2,3$, $j=1,\ldots,16$, a random vector whose entries are $X_1^j = $ Support, $X_2^j = $ Rejection, $X_3^j = $ Abstention and $n_j \in \{1200,1800\}$ is the sample size of survey $j$. These components represent exclusive selections (events) with probabilities

\[
P^j(X_1^j=x_1) = \theta_1^j, P^j(X_2^j=x_2) = \theta_2^j, P^j(X_3^j=x_3) = \theta_3^j, \ j=1,\ldots,16,
\]

where $\theta_1^j$, $\theta_2^j$ and $\theta_3^j$ are the percentages collected in Table \ref{TABLA1} for each survey $j$, $j=1,\ldots,16$. Thus, each random vector $X^j$ follows a multinomial (trinomial) probability distribution. Therefore, the probability that $X_1^j$ occurs $x_1$ times, $X_2^j$ occurs $x_2$ times and $X_3^j$ occurs $x_3$ times is given by

\[
P_{n_j}^j(x_1,x_2,x_3) = \frac{n_j!}{x_1! x_2! x_3!} (\theta_1^j)^{x_1} (\theta_2^j)^{x_2} (\theta_3^j)^{x_3}, \; j=1,\ldots, 16, 
\]

where $x_1+x_2+x_3=n_j$. The resulting trinomials for each Euskobarometro survey can be seen in the column 4 in Table \ref{TABLA2}.

\begin{table}[h]
\centering
\begin{small}
\begin{tabular}{|c|c|c|c|}
\hline
 & Survey  &Sample &	 Joint trinomial probability function	\\
 & dates	& size	&   \\
\hline
$j=1$	&	$t_{1}=$ May 05	&	$n_{1}=1800$	&	$P_{1800}^{1}(x_1,x_2,x_3) = \frac{1800!}{x_1! x_2! x_3!}0.02^{x_1}0.93^{x_2}0.05^{x_3}$	\\
$j=2$	&	$t_{2}=$ Nov 05	&	$n_{2}=1200$	&	$P_{1200}^{2}(x_1,x_2,x_3) = \frac{1200!}{x_1! x_2! x_3!}0.03^{x_1}0.93^{x_2}0.04^{x_3}$	\\
$j=3$	&	$t_{3}=$ May 06	&	$n_{3}=1800$	&	$P_{1800}^{3}(x_1,x_2,x_3) = \frac{1800!}{x_1! x_2! x_3!}0.03^{x_1}0.93^{x_2}0.04^{x_3}$	\\
$j=4$	&	$t_{4}=$ Nov 06	&	$n_{4}=1200$	&	$P_{1200}^{4}(x_1,x_2,x_3) = \frac{1200!}{x_1! x_2! x_3!}0.04^{x_1}0.86^{x_2}0.1^{x_3}$	\\
$j=5$	&	$t_{5}=$ May 07	&	$n_{5}=1200$	&	$P_{1200}^{5}(x_1,x_2,x_3) = \frac{1200!}{x_1! x_2! x_3!}0.02^{x_1}0.84^{x_2}0.14^{x_3}$	\\
$j=6$	&	$t_{6}=$ Nov 07	&	$n_{6}=1200$	&	$P_{1200}^{6}(x_1,x_2,x_3) = \frac{1200!}{x_1! x_2! x_3!}0.02^{x_1}0.9^{x_2}0.08^{x_3}$	\\
$j=7$	&	$t_{7}=$ May 08	&	$n_{7}=1800$	&	$P_{1800}^{7}(x_1,x_2,x_3) = \frac{1800!}{x_1! x_2! x_3!}0.03^{x_1}0.9^{x_2}0.07^{x_3}$	\\
$j=8$	&	$t_{8}=$ Nov 08	&	$n_{8}=1200$	&	$P_{1200}^{8}(x_1,x_2,x_3) = \frac{1200!}{x_1! x_2! x_3!}0.01^{x_1}0.93^{x_2}0.06^{x_3}$	\\
$j=9$	&	$t_{9}=$ May 09	&	$n_{9}=1200$	&	$P_{1200}^{9}(x_1,x_2,x_3) = \frac{1200!}{x_1! x_2! x_3!}0.04^{x_1}0.9^{x_2}0.06^{x_3}$	\\
$j=10$	&	$t_{10}=$ Nov 09	&	$n_{10}=1200$	&	$P_{1200}^{10}(x_1,x_2,x_3) = \frac{1200!}{x_1! x_2! x_3!}0.03^{x_1}0.89^{x_2}0.08^{x_3}$	\\
$j=11$	&	$t_{11}=$ May 10	&	$n_{11}=1200$	&	$P_{1200}^{11}(x_1,x_2,x_3) = \frac{1200!}{x_1! x_2! x_3!}0.03^{x_1}0.9^{x_2}0.07^{x_3}$	\\
$j=12$	&	$t_{12}=$ Nov 10	&	$n_{12}=1200$	&	$P_{1200}^{12}(x_1,x_2,x_3) = \frac{1200!}{x_1! x_2! x_3!}0.04^{x_1}0.88^{x_2}0.08^{x_3}$	\\
$j=13$	&	$t_{13}=$ May 11	&	$n_{13}=1200$	&	$P_{1200}^{13}(x_1,x_2,x_3) = \frac{1200!}{x_1! x_2! x_3!}0.04^{x_1}0.9^{x_2}0.06^{x_3}$	\\
$j=14$	&	$t_{14}=$ Nov 11	&	$n_{14}=1200$	&	$P_{1200}^{14}(x_1,x_2,x_3) = \frac{1200!}{x_1! x_2! x_3!}0.03^{x_1}0.89^{x_2}0.08^{x_3}$	\\
$j=15$	&	$t_{15}=$ May 12	&	$n_{15}=1200$	&	$P_{1200}^{15}(x_1,x_2,x_3) = \frac{1200!}{x_1! x_2! x_3!}0.05^{x_1}0.89^{x_2}0.06^{x_3}$	\\
$j=16$	&	$t_{16}=$ Nov 12	&	$n_{16}=1200$	&	$P_{1200}^{16}(x_1,x_2,x_3) = \frac{1200!}{x_1! x_2! x_3!}0.03^{x_1}0.92^{x_2}0.05^{x_3}$	\\
\hline 
\end{tabular}
\end{small} 
\caption{Data for probabilistic model estimation. Date, sample size and joint trinomial probability function of each survey.}
\label{TABLA2} 
\end{table}

\subsection{Probabilistic estimation}\label{3.2}
In this section, we are going to sample data survey for each survey, using the joint trinomial distribution set in Table \ref{TABLA2}. This will be done a high number of times ($10^4$ times) in order to generate a representative sample for each survey. Every time we sample data survey, we determine the model parameter estimations $\gamma_{12}$, $\gamma_{13}$, $\gamma_{23},$ using the Nelder-Mead optimization algorithm \cite{Nelder, Press} with goodness-of-fit $\chi^2$-test \cite{Groot}. The parameters with $p-$value less than $0.05$ will be rejected. The remainder will be sorted by $p-$value descending order. Selecting some of these model parameter vectors, we will be able to use the model outputs to provide a confidence band determined by the percentiles $2.5$ and $97.5$ ($95\%$ confidence interval) in each time instant. This $95\%$ model confidence band ($95\%$ MCB) is what we call \textit{probabilistic estimation}. Let us describe in detail the procedure.

\begin{figure}[h]
 \begin{center}
  \includegraphics[scale=0.45]{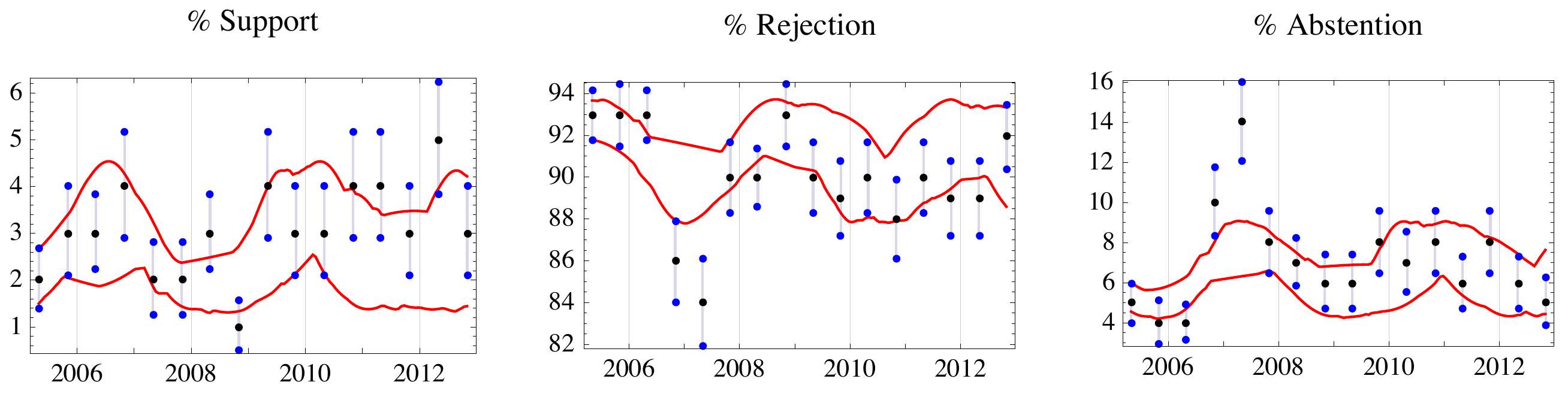}\\
  \caption{Probabilistic estimation. The vertical segments (error bars) correspond to the $95\%$ CI of the simulated data using multinomial distributions appearing in Table \ref{TABLA2}. The points in the middle of the segments are the mean values in Table \ref{TABLA1}. The continuous lines are the model $95\%$ MCB (probabilistic estimation) obtained with the described procedure. Note that most of the segments cross continuous lines determined by the model, capturing the data uncertainty. Only for Rejection and Abstention subpopulations in time instants Nov 2006 and May 2007 the uncertainty is not captured.} \label{bandas}
\end{center}
\end{figure}

\begin{enumerate}
\item Compute the quantiles $2.5$ and $97.5$ ($95\%$ CI) of each one of the joint multinomial distributions in Table \ref{TABLA2}, $j=1,2,\ldots,16$, for Support, Rejection and Abstention subpopulations sampling multinomials a hundred thousand times, obtaining

\begin{eqnarray}
Q_{2.5}^{support} & = & ( 1.39, 2.08, 2.22, 2.92, 1.25, 1.25, 2.22, 0.50, 2.92, 			\nonumber \\ 
                  &   &   2.08, 2.08, 2.92, 2.92, 2.08, 3.83, 2.08 ), 						\nonumber \\
Q_{97.5}^{support} & = & ( 2.67, 4.00, 3.78, 5.17, 2.83, 2.83, 3.83, 1.58, 5.17, 		\nonumber \\ 
                   &   &   4.00, 4.00, 5.17, 5.17, 4.00, 6.25, 4.00 ),   					\nonumber \\
Q_{2.5}^{reject} & = & ( 91.80, 91.50, 91.80, 84.00, 81.90, 88.20, 88.60, 91.50, 		\nonumber \\ 
                 &   &   88.20, 87.20, 88.20, 86.20, 88.20, 87.20, 87.20, 90.40 ), 		\nonumber \\
Q_{97.5}^{reject} & = & ( 94.20, 94.40, 94.20, 87.90, 86.10, 91.70, 91.40, 94.40, 		\nonumber \\
                  &   &   91.70, 90.70, 91.70, 89.80, 91.70, 90.70, 90.70, 93.50 ), 	\nonumber \\                    
Q_{2.5}^{abstention} & = & ( 4.00, 2.92, 3.11, 8.33, 12.10, 6.50, 5.83, 4.67, 4.67, 	\nonumber \\ 
                 &   &   6.50, 5.58, 6.50, 4.67, 6.50, 4.67, 3.83 ), 						\nonumber \\
Q_{97.5}^{abstention} & = & ( 6.00, 5.17, 4.94, 11.80, 16.00, 9.58, 8.22, 7.33, 7.33,  \nonumber \\
                  &   &   9.58, 8.50, 9.58, 7.33, 9.58, 7.42, 6.25 ).  						\nonumber   
\end{eqnarray}

The $95\%$ CI determined by the above percentiles (they can be seen in Figures \ref{bandas} and \ref{bandas2} as vertical segments (error bars)) constitute an approximation of the survey results. Moreover, these $95\%$ CI will be valuable to find the best probabilistic estimation.

\item Let us define the following function of the parameters $\gamma_{12}$, $\gamma_{13}$ and $\gamma_{23}$: 

\begin{itemize}
\item[A)] For given values of $\gamma_{12}$, $\gamma_{13}$ and $\gamma_{23}$ parameters, compute the model output in $t_1=$ May 2005, $t_2=$ Nov 2005, ..., $t_{15}=$ May 2012 and $t_{16}=$ Nov 2012 for the three subpopulations, Support, Rejection and Abstention.
\item[B)] Compare, for each subpopulation, the model output obtained in step (2A) to the data values we will sample in step (3A) using the $\chi^2$-test and obtain a $p-$value for each subpopulation.
\item[C)] Calculate the minimum $p-$value among the three above.
\end{itemize}

\item For $i$ = $1$ to $10^4$

\begin{itemize}
\item[A)] Sample another values of all the trinomial distributions in Table \ref{TABLA2}. Then, we will have one sample of $16$ surveys with percentages for Support, Rejection and Abstention populations from May 2005 until Nov 2012. Therefore, we will have a set of sampled data as in Table \ref{TABLA1}.
 
\item[B)] Find the model parameter values $\gamma_{12}^i$, $\gamma_{13}^i$ and $\gamma_{23}^i$ with the highest $p-$value (maximizing the function defined in steps (2A), (2B) and (2C)). To do that, Nelder-Mead optmization algorithm is used \cite{Nelder, Press} using as a goodness-of-fit the $\chi^2$-test.

\end{itemize}

\item Once the above process is completed, store the obtained parameter values and the $p-$value as the vector     

\[
 ( \gamma_{12}^i, \gamma_{13}^i, \gamma_{23}^i, p-\mbox{value}_i ), \ 1 \leq i \leq 10^4.
\]

\item Reject the model parameters with $p-$value less than $0.05$. In our case, $4990$ out of $10^4$ satisfy this restriction. Then, they are sorted by $p-$value descending order as follows,

\begin{equation}
 ( \gamma_{12}^i, \gamma_{13}^i, \gamma_{23}^i, p-\mbox{value}_i ), \ 1 \leq i \leq 4990. \label{BF}
\end{equation}

\item For $k$ = $2$ to $4990$

\begin{itemize}
\item[A)] Substitute into the model the parameters $( \gamma_{12}^j, \gamma_{13}^j, \gamma_{23}^j )$, for $j=1,2,\ldots, k$, and compute the model output in $t_1=$ May 2005, $t_2=$ Nov 2005, ..., $t_{15}=$ May 2012 and $t_{16}=$ Nov 2012.  

\begin{itemize}
\item[a1)] Take the $k$ model outputs for Support, Rejection and Abstention at time instant $t_1=$ May 2005 and calculate the corresponding quantiles $2.5$ and $97.5$ ($95\%$ CI).  
\item[a2)] Take the $k$ model outputs for Support, Rejection and Abstention at time instant $t_2=$ Nov 2005 and calculate the corresponding quantiles $2.5$ and $97.5$ ($95\%$ CI).  
\item $\cdots$
\item[a16)] Take the $k$ model outputs for Support, Rejection and Abstention at time instant $t_{16}=$ Nov 2012 and calculate the corresponding quantiles $2.5$ and $97.5$ ($95\%$ CI).  
\end{itemize}

\item[B)] Now, gather the calculated quantiles $2.5$ for Support, Rejection and Abstention subpopulations and store them sequentially on the vectors $S_{2.5}^k$, $R_{2.5}^k$ and $A_{2.5}^k$, respectively.
\item[C)] Gather the calculated quantiles $97.5$ for Support, Rejection and Abstention subpopulations and store them sequentially on the vectors $S_{97.5}^k$, $R_{97.5}^k$ and $A_{97.5}^k$, respectively.

\item[D)] Calculate the $p-$values using the $\chi^2$-test to datasets obtained in steps (1), (6B) and (6C) grouped in pairs as follows,

\begin{itemize}
\item[d1)] $Q_{2.5}^{support}$ and $S_{2.5}^k$,
\item[d2)] $Q_{97.5}^{support}$ and $S_{97.5}^k$,
\item[d3)] $Q_{2.5}^{reject}$ and $R_{2.5}^k$,
\item[d4)] $Q_{97.5}^{reject}$ and $R_{97.5}^k$,
\item[d5)] $Q_{2.5}^{abstention}$ and $A_{2.5}^k$,
\item[d6)] $Q_{97.5}^{abstention}$ and $A_{97.5}^k$.
\end{itemize}

Note that, in order to know the parameter values which allow us to define the $95\%$ MCB (probabilistic estimation), we compare percentil vectors obtained by the trinomial sampling to the obtained using the model outputs considering the $4990$ optimal values.
 
\item[E)] Calculate $m_k$ the minimum $p-$value among the six above and build the pair $(k, m_k)$.

\end{itemize}

\item Select the pair $(k, m_k)$ among the $4990$ with the maximum $m_k$.

\end{enumerate}

In our case, the obtained value is $k=77$ with $m_{77}=0.972991$ and consequently the $p$-values corresponding to percentiles $2.5$ and $97.5$ for each subpopulation are greater than $m_{77}$.

Now, we take the $k=77$ set of parameters obtained in the above procedure, compute the model output from $t_1=$ May 2005 to $t_{16}=$ Nov 2012, in jumps of $0.05$ and, in each point, we calculate the percentiles $2.5$ and $97.5$ for each subpopulation ($95\%$ MCB). The result (probabilistic estimation) is depicted in Figure \ref{bandas} as red continuous lines.

The vertical segments (error bars) correspond to the $95\%$ CI of the survey data simulated by multinomial distributions appearing in Table \ref{TABLA2}. The points in the middle of the segments are the mean values collected in Table \ref{TABLA1}. The continuous lines are the model $95\%$ MCB obtained from the model outputs of the first $k=77$ out of $4990$ sets of model parameters that best fit samples of the multinomial distributions in Table \ref{TABLA2}.

\subsection{Probabilistic estimation analysis}
The idea of the probabilistic estimation described in the previous section is to obtain $95\%$ model confidence interval bands (MCB) as close as possible, in the sense of $\chi^2$-test, to $95\%$ CI of the data distributions appearing in Table \ref{TABLA2} (vertical segments in Figure \ref{bandas}). This closeness depends on the model and on the data. In general, in Social Sciences and in particular in our case, these data are very sensitive to punctual events, not considered explicitly as model hypotheses, that in our model under study, may affect the attitude towards ETA. 

It is remarkable to note that the use of $\chi^2$-test in the procedure of the previous section to select the best fittings, allowed us to find $4990$ sets of model parameters for which the model estimation cannot be rejected as explanation of the data representing the studied phenomenon. In fact, we also could select the best (highest $p-$value) among all of them.  

In addition, looking at the graphics in Figure \ref{bandas}, we can see that almost all the vertical segments (error bars) cross at least a continuous line indicating that data uncertainty is captured by the model, in particular for Support subpopulation.

A mention deserves the Rejection and Abstention subpopulation graphics, where we can distinguish two parts. The first one, from May 2005 to May 2007, the probabilistic estimation intends to follow the data trajectory but the data uncertainty in Nov 2006 and May 2007 is not captured when sudden jumps appear. As we mentioned in Figure \ref{datos}, large jumps in the Rejection population correspond to large jumps in the Abstention population, in the opposite sense. We consider that the jumps in Nov 2006 and May 2007 are due to certain events that occurred from Sep 2006 to May 2007 as: increasing of vandalism acts from Sep 2006 to Dec 2006 linked with young left-wing nationalist groups; Barajas Airport Terminal 4 attack claimed by ETA (Dec 2006); in May 2007 local elections, the left-wing nationalist party EAE-ANV was allowed to present candidates in some villages and cities. In the second part, from Nov 2007 until Nov 2012, the continuous lines capture the data uncertainty.

Therefore, even though the estimation for Rejection and Abstention subpopulations do not capture the data uncertainty in two time instants, the three subpopulations capture the remainder and this leads us to consider the model and its probabilistic estimation appropriate to provide a prediction of the evolution of the population's attitude towards ETA over the next four years. 

\section{Probabilistic predictions over the next four years}
\begin{figure}[h]
 \begin{center}
  \includegraphics[scale=0.45]{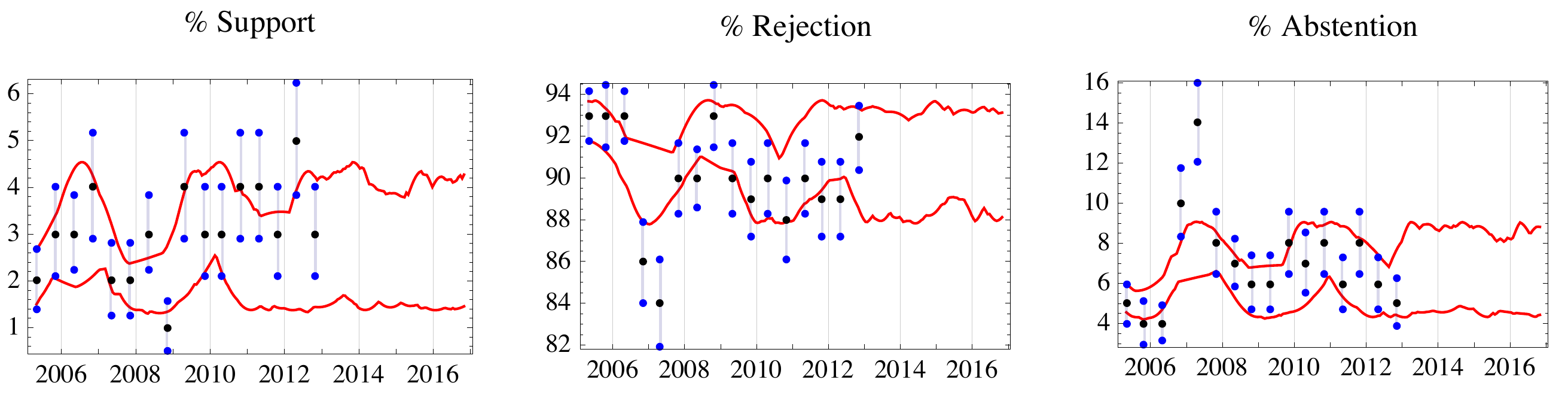}\\
  \caption{Probabilistic prediction. This picture is the  Figure \ref{bandas} including the predictions over the next four years as $95\%$ MCB (model continuous lines).} \label{bandas2}
\end{center}
\end{figure}

Now, taking the model and the $k=77$ set of parameters obtained in the probabilistic estimation, we are going to give the probabilistic prediction over the next four years by computing the model outputs from Nov 2012 to Nov 2016 and then, obtaining the $95\%$ MCB (model continuous lines). We plot the results graphically in Figure \ref{bandas2} and some numerical values in Table \ref{TABLA3}.

\begin{table}[h]
\centering
\begin{small}
\begin{tabular}{|c|c|c|c|c|c|c|}
\hline
Date    & \multicolumn{2}{c|}{Support} & \multicolumn{2}{c|}{Rejection} & \multicolumn{2}{c|}{Abstention} 	\\
		& Mean & $95\%$ CI & Mean & $95\%$ CI & Mean & $95\%$ CI \\
\hline
May 2013 & $  3.10$ & $[  1.54,  4.32]$ & $ 90.69$ & $[ 88.02, 93.38]$ & $  6.22$ & $[  4.52,  9.01]$ \\ 
Nov 2013 & $  2.98$ & $[  1.55,  4.54]$ & $ 90.28$ & $[ 88.11, 93.15]$ & $  6.74$ & $[  4.63,  8.87]$ \\ 
May 2014 & $  2.68$ & $[  1.41,  4.06]$ & $ 90.40$ & $[ 87.87, 92.90]$ & $  6.92$ & $[  4.63,  8.78]$ \\ 
Nov 2014 & $  2.43$ & $[  1.43,  3.88]$ & $ 90.86$ & $[ 88.21, 93.55]$ & $  6.71$ & $[  4.74,  8.90]$ \\ 
May 2015 & $  2.44$ & $[  1.47,  3.84]$ & $ 91.23$ & $[ 89.03, 93.41]$ & $  6.34$ & $[  4.33,  8.66]$ \\ 
Nov 2015 & $  2.62$ & $[  1.42,  4.26]$ & $ 91.22$ & $[ 88.76, 93.25]$ & $  6.17$ & $[  4.60,  8.08]$ \\ 
May 2016 & $  2.72$ & $[  1.42,  4.17]$ & $ 91.01$ & $[ 88.56, 93.39]$ & $  6.27$ & $[  4.49,  8.99]$ \\ 
Nov 2016 & $  2.72$ & $[  1.46,  4.28]$ & $ 90.87$ & $[ 88.13, 93.13]$ & $  6.41$ & $[  4.43,  8.82]$ \\ 
\hline 
\end{tabular} 
\end{small}
\caption{Mean and $95\%$ confidence interval predictions for the coming eight Euskobarometro surveys.}
\label{TABLA3} 
\end{table}

Figure \ref{bandas2} and Table \ref{TABLA3} show us that the attitude towards ETA of the population living in the Basque Country will remain fairly stable over the next four years.

\subsection{Robustness of the presented method}
Note that if we run the described procedure again, taking into account that the multinomial sampling is random, we may obtain a different value of $k$, however, the corresponding $m_k$ will be very similar. In fact, we did it twice more obtaining $k=129$ and $k=84$ with $m_k=0.9624$ and $m_k=0.966348$, respectively. The probabilistic estimations in these two new cases are given in Tables \ref{TABLA4} and \ref{TABLA5}. We can see that the predictions were very similar. This shows the robustness of the proposed method.

\begin{table}[h]
\centering
\begin{small}
\begin{tabular}{|c|c|c|c|c|c|c|}
\hline
Date    & \multicolumn{2}{c|}{Support} & \multicolumn{2}{c|}{Rejection} & \multicolumn{2}{c|}{Abstention} 	\\
		& Mean & $95\%$ CI & Mean & $95\%$ CI & Mean & $95\%$ CI \\
\hline
May 2013 & $  3.08$ & $[  1.48,  4.76]$ & $ 91.02$ & $[ 88.40, 93.44]$ & $  5.90$ & $[  4.42,  8.33]$ \\ 
Nov 2013 & $  3.09$ & $[  1.59,  4.61]$ & $ 90.41$ & $[ 87.90, 93.33]$ & $  6.50$ & $[  4.35,  8.97]$ \\ 
May 2014 & $  2.86$ & $[  1.39,  4.52]$ & $ 90.30$ & $[ 88.05, 92.89]$ & $  6.84$ & $[  4.43,  8.89]$ \\ 
Nov 2014 & $  2.59$ & $[  1.48,  4.22]$ & $ 90.64$ & $[ 88.50, 93.28]$ & $  6.77$ & $[  4.15,  8.71]$ \\ 
May 2015 & $  2.45$ & $[  1.46,  3.86]$ & $ 91.10$ & $[ 88.54, 93.21]$ & $  6.45$ & $[  3.89,  8.53]$ \\ 
Nov 2015 & $  2.51$ & $[  1.56,  3.93]$ & $ 91.35$ & $[ 88.98, 93.31]$ & $  6.14$ & $[  3.67,  8.41]$ \\ 
May 2016 & $  2.67$ & $[  1.47,  4.32]$ & $ 91.28$ & $[ 88.79, 93.33]$ & $  6.05$ & $[  3.56,  8.29]$ \\ 
Nov 2016 & $  2.75$ & $[  1.31,  4.40]$ & $ 91.06$ & $[ 88.09, 93.19]$ & $  6.19$ & $[  3.74,  8.75]$ \\ 
\hline 
\end{tabular} 
\end{small}
\caption{Mean and $95\%$ confidence interval predictions for the coming eight Euskobarometro surveys for the second procedure execution, $k=129$ and $m_k=0.9624$.}
\label{TABLA4} 
\end{table}

\begin{table}[h]
\centering
\begin{small}
\begin{tabular}{|c|c|c|c|c|c|c|}
\hline
Date    & \multicolumn{2}{c|}{Support} & \multicolumn{2}{c|}{Rejection} & \multicolumn{2}{c|}{Abstention} 	\\
		& Mean & $95\%$ CI & Mean & $95\%$ CI & Mean & $95\%$ CI \\
\hline
May 2013 & $  3.13$ & $[  1.73,  4.79]$ & $ 90.89$ & $[ 88.29, 93.22]$ & $  5.98$ & $[  4.47,  8.32]$ \\ 
Nov 2013 & $  3.08$ & $[  1.79,  4.59]$ & $ 90.34$ & $[ 87.87, 92.99]$ & $  6.57$ & $[  4.38,  8.81]$ \\ 
May 2014 & $  2.85$ & $[  1.47,  4.61]$ & $ 90.31$ & $[ 88.17, 92.28]$ & $  6.84$ & $[  4.79,  8.89]$ \\ 
Nov 2014 & $  2.56$ & $[  1.47,  4.23]$ & $ 90.66$ & $[ 88.11, 92.88]$ & $  6.78$ & $[  4.79,  8.67]$ \\ 
May 2015 & $  2.39$ & $[  1.50,  3.66]$ & $ 91.12$ & $[ 88.65, 93.34]$ & $  6.50$ & $[  4.27,  8.51]$ \\ 
Nov 2015 & $  2.48$ & $[  1.54,  3.84]$ & $ 91.34$ & $[ 89.30, 93.05]$ & $  6.18$ & $[  4.40,  8.37]$ \\ 
May 2016 & $  2.67$ & $[  1.47,  4.26]$ & $ 91.17$ & $[ 88.63, 93.08]$ & $  6.15$ & $[  4.71,  8.09]$ \\ 
Nov 2016 & $  2.71$ & $[  1.43,  4.35]$ & $ 90.94$ & $[ 87.93, 93.28]$ & $  6.35$ & $[  4.18,  8.69]$ \\ 
\hline 
\end{tabular} 
\end{small}
\caption{Mean and $95\%$ confidence interval predictions for the coming eight Euskobarometro surveys for the third procedure execution, $k=84$ and $m_k=0.966348$.}
\label{TABLA5} 
\end{table}

Also, we should say that last June 27th, 2013 was published the Euskobarometro of May 2013 with $1200$ interviews and values given in Table \ref{TABLA6}. The $95\%$ confidence intervals of this last Euskobarometro were calculated as the Step 1 of the procedure described in Section \ref{3.2}.

\begin{table}[h]
\centering
\begin{small}
\begin{tabular}{|c|c|c|c|c|c|c|}
\hline
Date    & \multicolumn{2}{c|}{Support} & \multicolumn{2}{c|}{Rejection} & \multicolumn{2}{c|}{Abstention} 	\\
		& Mean & $95\%$ CI & Mean & $95\%$ CI & Mean & $95\%$ CI \\
\hline
May 2013 & $3$ & $[ 2.08,  4.00]$ & $89$ & $[ 87.17, 90.75]$ & $8$ & $[  6.50,  9.58]$ \\ 
\hline 
\end{tabular} 
\end{small}
\caption{Mean and $95\%$ confidence interval of the Euskobarometro corresponding to May 2013.}
\label{TABLA6} 
\end{table}

Comparing data in Table \ref{TABLA6} to results in Tables \ref{TABLA3}, \ref{TABLA4} and \ref{TABLA5}, we can see that the data uncertainty in Euskobarometro May 2013 is captured by our predictions in the three tables.

\section{Conclusion}
In this paper, it is presented a computational technique to deal with uncertainty (in parameter estimation and output predictions) in dynamic social models based on systems of differential equations. This technique takes data from surveys to introduce the uncertainty into the model from the very beginning and returns $95\%$ model confidence interval bands that capture the data uncertainty and predict what will happen over the next future. 
  
In order to present the possibilities of this technique, it is applied to a mathematical model to study the evolution dynamics of the attitude of Basque population towards ETA. Once the model is stated, we determine a probabilistic estimation in order to find out if the model captures the Euskobarometro data evolution. We observe that the model captures the data uncertainty only partially from May 2005 to May 2007, but from Nov 2007 the probabilistic estimation improves perceptibly. Anyway, the model estimation is non-rejectable using the $\chi^2$-test. Then, we provide a probabilistic prediction of the attitude towards ETA of the population of the Basque Country over the next four years. The predicted results point out that, if the political scenario does not change, the current situation will remain fairly stable.

Additionally, some benefits that can be obtained with this approach are:

\begin{itemize}
\item If we consider the model parameters as random variables, the technique presented here as probabilistic estimation allows the estimation of samples of these model parameters (Steps 3, 4 and 5 of the procedure described in Section \ref{3.2}). This fact is of paramount interest because one of main challenges in modelling real problems using random differential equations is to determine the distribution function of model parameters. Therefore, if we use the probabilistic estimation to obtain some samples of the model parameters, in our case $k=77$ parameter samples, we can use these samples and statistical hypothesis testing or kernel functions in order to find distribution functions of the model parameters. 
\item Other aspect that should be mentioned and it could be interesting for survey prediction estimations is the fact that Table \ref{TABLA3} (\ref{TABLA4} and \ref{TABLA5}) may be considered as an estimation of the results of the coming Euskobarometro surveys (mean and $95\%$ confidence interval). This idea may be applied to this and other type of surveys where a reliable underlying dynamic model can be built. As a consequence, some surveys may not be carried out with the corresponding saving of money. Therefore, we consider that this approach may be an interesting tool for social behavior studies.
\end{itemize}

\end{document}